\documentclass[11pt]{article}
\usepackage[final]{acl}

\usepackage{times}
\usepackage{latexsym}
\usepackage[T1]{fontenc}
\usepackage[utf8]{inputenc}
\usepackage{microtype}
\usepackage{inconsolata}

\usepackage{graphicx}
\usepackage{booktabs}
\usepackage{multirow}
\usepackage{amsmath}

\title{DialectSentEval 2026: Arabic Dialect Sentiment Analysis and Swapping Shared Task}

\author{
  \textbf{Saad Ezzini\textsuperscript{1,*}},
  \textbf{Shadi Abudalfa\textsuperscript{2,3}},
  \textbf{Maram Alharbi\textsuperscript{4}},
  \textbf{Salmane Chafik\textsuperscript{5}},\\
  \textbf{Hind Alatawi\textsuperscript{1}},
  \textbf{Mo El-Haj\textsuperscript{4,6}},
  \textbf{Ahmed Abdelali\textsuperscript{7}},
  \textbf{Osamah Alnahari\textsuperscript{1}},
  \textbf{Salima Lamsiyah\textsuperscript{8}}
\\
  \textsuperscript{1}King Fahd University of Petroleum and Minerals, KSA,
  \textsuperscript{2}Onaizah Colleges, KSA,\\
  \textsuperscript{3}University College of Applied Sciences, Palestine,
  \textsuperscript{4}Lancaster University, UK,\\
  \textsuperscript{5}UM6P, Morocco,
  \textsuperscript{6}VinUniversity, Vietnam,
  \textsuperscript{7}El Technology, Qatar,\\
  \textsuperscript{8}University of Luxembourg, Luxembourg
\\
  {
    \textbf{*Correspondence:} \href{mailto:saad.ezzini@kfupm.edu.sa}{saad.ezzini@kfupm.edu.sa}
  }
}

\begin{document}
\maketitle

\begin{abstract}
Sentiment analysis is a fundamental problem in Natural Language Processing (NLP). Standard sentiment classification for the Arabic language remains challenging due to the high volume of dialectal Arabic. To advance research in this area, this paper proposes the Shared Task on Sentiment Analysis and Swapping in Arabic Dialects (DialectSentEval), hosted with the Arabic Natural Language Processing Conference (ArabicNLP 2026). This shared task consists of two subtasks: Subtask 1 focuses on multi-class and multi-dialect sentiment analysis, requiring models to identify sentiment polarity across various Arabic dialects. Subtask 2 introduces a generative task for Arabic sentiment swap, challenging models to invert sentiment polarity while preserving core semantics. In this overview paper, we present the motivation, dataset creation, and summarize the main findings from participating models.
\end{abstract}

\section{Introduction}

Sentiment analysis remains a key task in Natural Language Processing (NLP), with critical real-world applications ranging from social media monitoring to customer feedback analysis. While significant progress has been made for many languages, and multiple types have been defined (document, sentence, and feature/aspect level), the Arabic language presents unique challenges due to its rich morphology and extensive dialectal variation. Most user-generated content, the primary source for sentiment data, is written in dialectal Arabic, which is often under-resourced and differs significantly from Modern Standard Arabic (MSA).

To address these challenges and foster innovation in Arabic NLP, we propose a novel Shared Task on Sentiment Analysis and Swapping in Arabic Dialects (DialectSentEval), hosted with the Arabic Natural Language Processing Conference (ArabicNLP 2026)\footnote{\url{ https://ezzini.github.io/DialectSentEval/}}. This task is designed to move beyond standard sentiment classification by evaluating both the understanding and generation of sentiment in diverse Arabic contexts. DialectSentEval comprises two distinct but complementary subtasks:

\begin{itemize}
    \item \textbf{Subtask 1: Arabic Dialect Sentiment Analysis}: A multi-class classification task focused on identifying the sentiment (positive, negative, or neutral) of texts written in various Arabic dialects. This subtask emphasizes the need for robust models that can handle the lexical and syntactic diversity of the Arab world.
    \item \textbf{Subtask 2: Arabic Sentiment Swap}: A generative task where participants' systems must rewrite a given sentence to invert its sentiment polarity (e.g., positive to negative) while preserving its core meaning. This challenges models to go beyond surface-level cues and demonstrate a deeper grasp of semantics and syntax.
\end{itemize}

\section{Motivation}

The motivation for DialectSentEval is grounded in both linguistic and methodological gaps in current Arabic NLP research.

First, despite sustained interest in sentiment analysis for Arabic, there remains a clear shortage of robust benchmarks that adequately capture dialectal diversity. Earlier shared tasks and evaluation campaigns \citep{ElBeltagy2017, Rosenthal2017} have advanced sentiment classification for Arabic, yet most resources are either limited to MSA or focus on a narrow set of dialects and domains \cite{elhaj2026tarab,elhaj2020habibi}. In practice, however, the majority of user-generated content, including reviews, social media posts, and online discussions, is written in dialectal Arabic. These dialects exhibit substantial lexical, morphological, and syntactic variation, which often leads to performance degradation when models trained on MSA are applied to real-world data. Prior work has highlighted the fragmentation of Arabic sentiment resources and the difficulty of transferring models across dialects \citep{nwesri-etal-2025-sentiment,alwakid2022muldasa}, reinforcing the need for more representative evaluation benchmarks. By introducing a balanced, multi-dialect benchmark informed by recent benchmarking efforts \citep{alharbi-etal-2025-evaluating-large}, DialectSentEval seeks to promote the development of sentiment models that generalise across dialects and better reflect authentic language use in the Arab world.

Second, the task aims to extend evaluation beyond classification into controlled text generation. While sentiment classification has reached a relatively mature stage in Arabic NLP, generative sentiment manipulation remains underexplored \citep{abudalfa2025emerging}. Sentiment swapping, framed as a form of style transfer, requires systems to invert polarity while preserving core propositional content. This problem has been investigated in English \citep{Shen2017, li-etal-2018-delete}, where style transfer research has demonstrated the complexity of disentangling content from sentiment and other stylistic attributes. Comparable resources and systematic evaluation frameworks are scarce for Arabic, limiting progress in controllable generation \cite{abudalfa2024arabic,abdu2025empirical}. The limited availability of high-quality parallel datasets has constrained progress in this area. By introducing a dedicated sentiment swap subtask, DialectSentEval encourages research on semantic preservation, polarity control, and evaluation methodologies for Arabic natural language generation. Such capabilities have practical implications for data augmentation, controllable dialogue systems, and adaptive content generation, particularly in low-resource and dialectally diverse settings.

By combining a dialect-aware classification task with a challenging generative task, DialectSentEval offers a unified benchmark for both sentiment understanding and sentiment manipulation in Arabic. This dual focus is intended to stimulate research that moves beyond surface-level polarity detection towards more semantically grounded and controllable language technologies for Arabic.

\section{Data Collection and Creation}

\begin{figure*}
    \centering
    \includegraphics[width=1.8\columnwidth]{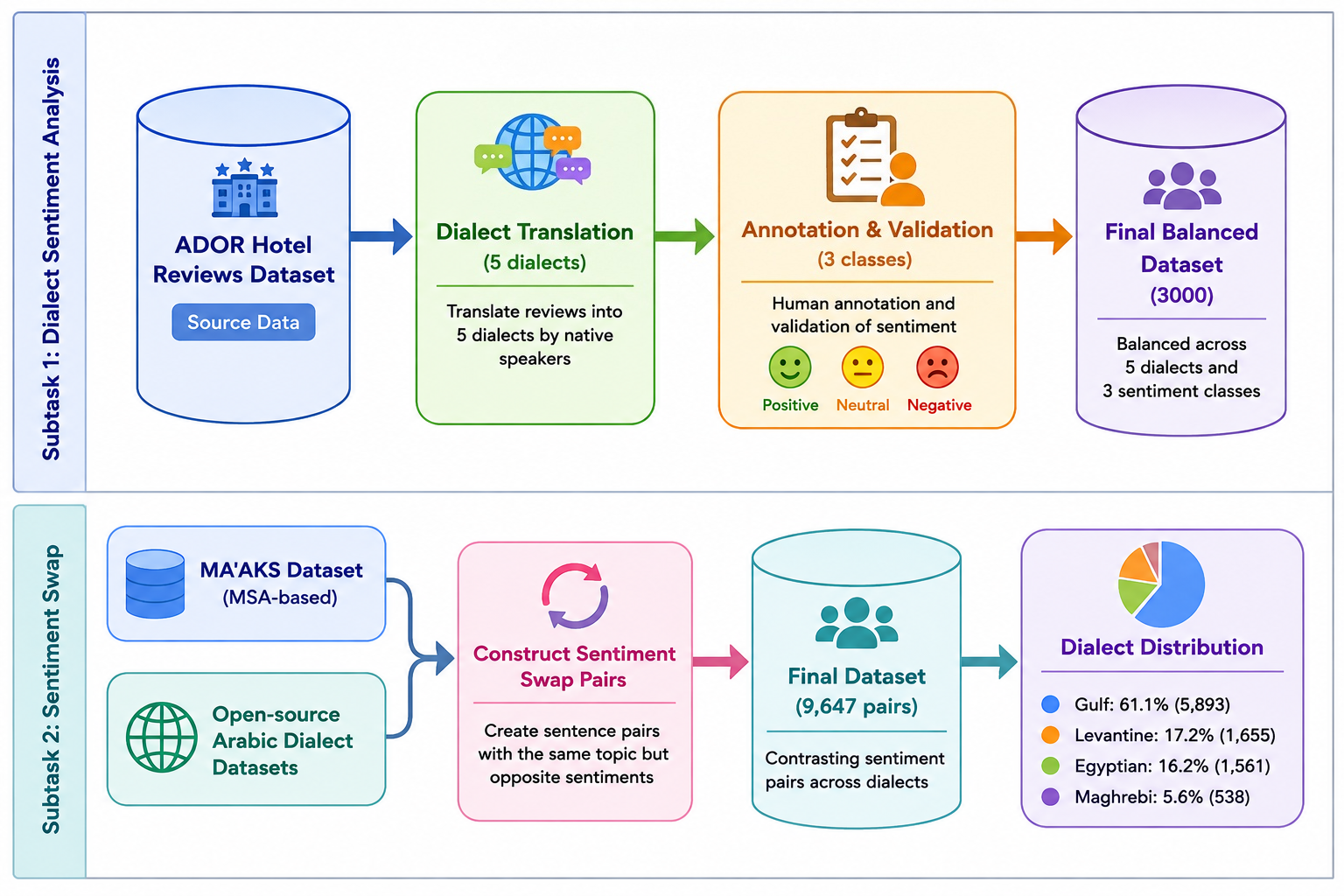}
    \caption{Data creation process for both subtasks.}
    \label{fig:colorfull_Final2}
\end{figure*}

The datasets for both subtasks were collected and annotated prior to the task launch. Figure \ref{fig:colorfull_Final2} illustrates the dataset creation process for both subtasks.

\subsection{Subtask 1: Dialect Sentiment Analysis}

The dataset for this task, Multi-Dialect-Sent (MDS-3), currently consists of 3,000 sentences annotated for three-class sentiment (positive, negative, neutral). The dataset is balanced across five major dialects: Moroccan, Egyptian, Lebanese, Jordanian, and Saudi. The data was sourced from hotel reviews from the ADOR dataset \citep{alharbi2025ador}, then translated and annotated by native speakers of each dialect. The sentences are parallel across dialects, with balanced sentiment representation across all included dialects. An initial version of the dataset was verified in recently organised shared tasks Ahasis \citep{alharbi2025ahasis} and AraSentEval \citep{ezzini2026arasenteval}. 

The dataset was split into 80\% for training and 20\% for testing. The split was performed based on sentence IDs to ensure that parallel sentences remain in the same set, to prevent potential data leakage between the training and test sets.

\subsection{Subtask 2: Sentiment Swap}

This subtask draws on a dataset built as an extension of MA'AKS \citep{mughaus2026ma}, comprising a total of 9,647 sentence-pair instances. Each pair centers on a shared topic but conveys contrasting sentiment orientations, with one member expressing a positive stance and the other a negative one. Of the full collection, 5,331 instances (55.3\%) carry a Negative label, while the remaining 4,316 (44.7\%) are tagged Positive. 

Whereas the original MA'AKS resource was built predominantly around MSA, the present corpus is drawn entirely from regional Arabic dialects and spans four principal dialectal families: Gulf (5,893 pairs, 61.1\% of the total), Levantine (1,655 pairs, 17.2\%), Egyptian (1,561 pairs, 16.2\%), and Maghrebi (538 pairs, 5.6\%).

To ensure reliability, every pair passed through a rigorous validation process assessing fluency, semantic fidelity, and the correctness of the sentiment reversal. The corpus comes with fixed training, validation, and test partitions, requiring models to capture both the meaning and the sentiment structure of each sentence. These partitions consist of 6,752 pairs for training, 964 for validation, and 1,931 for testing.

\section{Task Description and Evaluation}

In this section, we describe in detail the subtasks that define our shared task. Participants were allowed to participate in either one or both subtasks. The official evaluation was conducted on the \textbf{CodaBench} platform, an open-source platform designed to organize Artificial Intelligence benchmarks by defining tasks, including datasets, submission pipelines, and evaluation metrics. It provides an automated and transparent framework for evaluating system submissions.

\subsection{Subtask 1: Arabic Dialect Sentiment Analysis}

In this subtask, participants are required to build a system that classifies the sentiment expressed in sentences written in five different and linguistically challenging Arabic dialects: Moroccan, Egyptian, Lebanese, Jordanian, and Saudi. The goal of this task is to evaluate the ability of models to handle dialectal variation in Arabic while accurately identifying the sentiment conveyed in each sentence.

Formally, the objective is to design a sentence-level classifier \(C\) such that, given an input sentence \(x\), the system predicts a sentiment label \(S\), where:

\[
S = C(x), \quad S \in \{ \text{Positive}, \text{Neutral}, \text{Negative} \}.
\]

\begin{itemize}
    \item \textbf{Input}: A sentence written in Arabic using one of the target dialects (Moroccan, Egyptian, Lebanese, Jordanian, or Saudi).
    
    \item \textbf{Output}: A sentiment label selected from the set \{Positive, Neutral, Negative\}.
    
    \item \textbf{Evaluation Metric}: Systems are evaluated using the \textbf{Macro F1-score}. This metric is chosen because it gives equal importance to all classes and is therefore more suitable in scenarios where class distributions may be imbalanced.

\end{itemize}

\subsection{Subtask 2: Arabic Sentiment Swap}

In the second subtask, participants are expected to design models able to generate a new sentence that expresses the opposite sentiment of a given sentence while preserving its original semantic content. The input sentences are written in one of several Arabic dialect groups: Maghreb, Egyptian, Gulf, and Levantine. This task evaluates a model’s ability to perform controlled sentiment transformation while maintaining semantic fidelity.

Formally, the objective is to train a model \(M\) capable of performing sentiment-controlled text generation. Given an input sentence \(x\) expressing a sentiment \(S\), the model should generate a new sentence \(x'\) that conveys the opposite sentiment \(S'\), while preserving the core meaning and topic of the original sentence, such that:

\[
x' = M(x), \quad S' = \text{Opposite}(S)
\]

where \(S, S' \in \{\text{Positive}, \text{Negative}\}\) and \(S' \neq S\).

\begin{itemize}
    \item \textbf{Input}: An Arabic sentence along with its original sentiment polarity.
    
    \item \textbf{Output}: A rewritten Arabic sentence that preserves the main meaning of the input while expressing the opposite sentiment polarity.
    
    \item \textbf{Evaluation Metrics}: The generated outputs are evaluated using both sentiment accuracy and text similarity metrics to ensure correct sentiment transfer while maintaining semantic consistency.
    \begin{itemize}
        \item \textbf{Sentiment Style Accuracy}: A pre-trained sentiment classifier from CamelLab\footnote{CAMeL-Lab/bert-base-arabic-camelbert-da-sentiment} is used to measure the percentage of generated sentences that correctly express the target polarity.
        
        \item \textbf{BLEU and chrF}: These metrics are used to evaluate the degree of lexical and character-level similarity between the generated sentence and the reference sentence, providing an indication of semantic preservation.
    \end{itemize}
\end{itemize}

\section{Participant Systems}

Across both subtasks, a total of 15 unique teams participated in the competition, with 16 system description papers submitted. Below, we describe the methodologies and system architectures proposed by the participating teams, categorized by subtask.

\subsection{Subtask 1: Arabic Dialect Sentiment Analysis}

Participating teams for Subtask 1 investigated a diverse range of pretrained transformer architectures, ensembling strategies, loss function formulations, and domain-adaptation techniques to tackle multi-dialect variation and zero-shot dialect transfer.

\begin{itemize}
    \item \textbf{ALEXIS} (Username: \textit{momen\_ibrahim-870387}) \citep{ibrahim-etal-2026-alexis-s1}: Ranked \textbf{1st} overall with a Macro F1-score of \textbf{0.8667}. The team proposed an adversarial training and iterative self-training pipeline built on top of MARBERTv2. Specifically, they utilized the Fast Gradient Method (FGM) during fine-tuning to regularize the feature space against dialectal perturbations. To address zero-shot generalization on unlabelled dialects (notably Lebanese), they implemented a confidence-thresholded pseudo-labeling strategy that iteratively incorporated high-confidence test predictions back into training.
    
    \item \textbf{Axiom} (Username: \textit{ajwad\_hossain-867052}) \citep{hossain-2026-axiom}: Ranked \textbf{2nd} with a Macro F1-score of \textbf{0.8210}. They developed a dialect-aware ensemble framework combining MARBERTv2 and AraBERTv2. Their system utilized Stratified 5-Fold Cross-Validation, text normalization tailored to dialectal orthography, and an automated soft-voting probability aggregation scheme.
    
    \item \textbf{PTUK-NLP} (Username: \textit{tasneemduridi-857223}) \citep{duridi-etal-2026-ptuk}: Ranked \textbf{3rd} with a Macro F1-score of \textbf{0.8174}. The team fine-tuned MARBERT with domain-specific cleaning pipelines. They demonstrated that representations learned from geographically and linguistically proximate Levantine variants (Jordanian) transfer effectively in a zero-shot fashion to Lebanese. They also provided an extensive qualitative error analysis identifying mixed-polarity reviews as the primary source of classification errors.
    
    \item \textbf{NU-SentEval} (Username: \textit{habiba\_ayman-867381}) \citep{ayman-etal-2026-nu}: Ranked \textbf{4th} with a Macro F1-score of \textbf{0.8140}. They developed a stacked model blending predictions from MARBERTv2, AraBERTv02, and QARiB. Crucially, they integrated Focal Loss during training to mitigate class imbalance and focus gradient updates on hard-to-classify neutral examples.
    
    \item \textbf{Gradient Descenders} (Username: \textit{normalman159-857755}) \citep{nhan-etal-2026-gradient}: Ranked \textbf{5th} with a Macro F1-score of \textbf{0.8010}. The team performed an empirical comparison across multiple Arabic encoders (MARBERT, ARBERT, and AraBERT), concluding that fine-tuned MARBERTv2 combined with class-weighted cross-entropy loss and linear learning-rate warm-up yielded optimal multi-dialect performance.
    
    \item \textbf{LahjaSensei} (Username: \textit{ahmadalmoustafa-856400}) \citep{almoustafa-2026-lahjasensei}: Ranked \textbf{6th} with a Macro F1-score of \textbf{0.7977}. They utilized fine-tuned transformer backbones accompanied by dialect-specific token normalization and prompt-based tuning.
    
    \item \textbf{NAMAA Community} (Username: \textit{mona\_-871069} / \textit{hsnhb7}) \citep{abdelazim-etal-2026-namaa}: Ranked \textbf{7th} with a Macro F1-score of \textbf{0.7953}. They constructed an ensemble merging MARBERTv2, AraBERTv2, and AraModernBERT, using dialect classification heads to dynamically weigh base-model outputs.
    
    \item \textbf{faisalm93} (Username: \textit{faisalm93-869067}) \citep{adam-2026-faisalm93}: Ranked \textbf{8th} with a Macro F1-score of \textbf{0.7939}. The author developed a 5-fold cross-validated MARBERTv2 bagging ensemble utilizing soft voting across fold models, Cosine Annealing learning rate schedules, and layer-wise learning rate decay.
    
    \item \textbf{Lahja} (Username: \textit{shadenalzeer-867546}) \citep{alzeer-2026-lahja}: Ranked \textbf{10th} with a Macro F1-score of \textbf{0.7894}. They implemented a multi-seed ensembled MARBERT architecture designed to stabilize zero-shot transfer for the unrepresented Lebanese dialect by exploiting cross-dialect semantic regularities.
    
    \item \textbf{Northwestern} (Username: \textit{monmon2000}) \citep{shlkamy-zaghouani-2026-northwestern}: Explored a comparative study of MARBERTv2 and AraBERT under various dialectal splits, demonstrating that dialect-specific pretraining provides significant robustness gains over general MSA models.
\end{itemize}

\subsection{Subtask 2: Arabic Sentiment Swap}

Subtask 2 saw participating systems leverage generative large language models (LLMs), parameter-efficient adapters, candidate reranking pipelines, and preference alignment techniques to tackle the complex dual objective of sentiment polarity inversion and semantic fidelity.

\begin{itemize}
    \item \textbf{ALEXIS} (Username: \textit{momen\_ibrahim-870527}) \citep{ibrahim-etal-2026-alexis-s2}: Ranked \textbf{1st} across all evaluation metrics, achieving a Sentiment Style Accuracy of \textbf{0.9581}, a BLEU score of \textbf{56.91}, and a chrF of \textbf{72.18}. They introduced a multi-stage framework combining: (1) dialect-aware Retrieval-Augmented Generation (RAG) using semantic similarity over training pairs, (2) Direct Preference Optimization (DPO) to align open-weight generative backbones towards clean polarity transfer, and (3) a classifier-guided candidate reranker that filters out candidates failing the target sentiment direction.
    
    \item \textbf{AKARROUCH} (Username: \textit{akarrouch-871040}) \citep{akarrouch-2026-akarrouch}: Ranked \textbf{2nd} with a Sentiment Style Accuracy of \textbf{0.7851}, BLEU of \textbf{49.82}, and chrF of \textbf{70.12}. The system employed a generate-and-rerank architecture where an instruction-tuned LLM produced multiple candidate rewrites via temperature sampling, followed by a fine-tuned sentiment classifier to select the candidate maximizing target polarity probability.
    
    \item \textbf{LahjaSensei} (Username: \textit{ahmadalmoustafa-865064}) \citep{almoustafa-2026-lahjasensei}: Ranked \textbf{3rd} with a Sentiment Style Accuracy of \textbf{0.7789}, BLEU of \textbf{41.48}, and chrF of \textbf{64.45}. They proposed few-shot in-context prompting combined with Model Merging via Spherical Linear Interpolation (SLERP) across fine-tuned dialect checkpoints.
    
    \item \textbf{NAMAA} (Username: \textit{mona\_-871054} / \textit{hsnhb7-868552}) \citep{barmandah-etal-2026-namaa}: Ranked \textbf{4th} with a Sentiment Style Accuracy of \textbf{0.7597}, BLEU of \textbf{51.58}, and chrF of \textbf{71.02}. The team introduced a cross-backbone ensembling framework combining candidate hypotheses from Qwen2.5, Llama-3, and AraT5, reranked by an auxiliary sentiment verification classifier.
    
    \item \textbf{Sabaa} (Username: \textit{mohamedsabaa-855812}) \citep{sabaa-etal-2026-sabaa}: Ranked \textbf{6th} with a Sentiment Style Accuracy of \textbf{0.7447}, BLEU of \textbf{49.07}, and chrF of \textbf{69.33}. They utilized fine-tuned AraT5 sequence-to-sequence models combined with an external MARBERT-based sentiment classifier to rerank beam-search hypotheses.
    
    \item \textbf{NU-SentEval} (Username: \textit{areejkhaled-868894}) \citep{tamer-etal-2026-nu}: Ranked \textbf{8th} with a Sentiment Style Accuracy of \textbf{0.6898}, achieving a top-tier chrF of \textbf{72.61} and BLEU of \textbf{54.63}. They fine-tuned AraT5-base models conditioned on explicit dialect and polarity transformation prefix tokens.
    
    \item \textbf{NileNLP} (Username: \textit{mohammed\_bahgat-863352} / \textit{ehabashraf-863355}) \citep{bahgat-etal-2026-nilenlp}: Ranked \textbf{9th--10th} with a Sentiment Style Accuracy of \textbf{0.6732}, BLEU of \textbf{41.38}, and chrF of \textbf{65.84}. They developed \textit{SWAPPER}, a parameter-efficient adapter framework applying Low-Rank Adaptation (LoRA) to generative Arabic models.
\end{itemize}

\section{Results and Analysis}

\subsection{Subtask 1: Arabic Dialect Sentiment Analysis}

Table~\ref{tab:results_task1} reports the official evaluation results on the test set for Subtask 1, ranked by Macro F1-score.

\begin{table*}[ht]
\centering
\small
\begin{tabular}{rllc}
\toprule
\textbf{\#} & \textbf{Team Name} & \textbf{CodaBench Username} & \textbf{Macro F1} \\
\midrule
1 & \textbf{ALEXIS} & \textit{momen\_ibrahim-870387} & \textbf{0.8667} \\
2 & \textbf{Axiom} & \textit{ajwad\_hossain-867052} & 0.8210 \\
3 & \textbf{PTUK-NLP} & \textit{tasneemduridi-857223} & 0.8174 \\
4 & \textbf{NU-SentEval} & \textit{habiba\_ayman-867381} & 0.8140 \\
5 & \textbf{Gradient Descenders} & \textit{normalman159-857755} & 0.8010 \\
6 & \textbf{LahjaSensei} & \textit{ahmadalmoustafa-856400} & 0.7977 \\
7 & \textbf{NAMAA Community} & \textit{mona\_-871069} & 0.7953 \\
8 & \textbf{faisalm93} & \textit{faisalm93-869067} & 0.7939 \\
9 & -- & \textit{hoanguitcs-866914} & 0.7936 \\
10 & \textbf{Lahja} & \textit{shadenalzeer-867546} & 0.7894 \\
11 & -- & \textit{hh03112005-869469} & 0.7859 \\
12 & -- & \textit{23020073-870754} & 0.7722 \\
\bottomrule
\end{tabular}
\caption{Official evaluation results for Subtask 1: Arabic Dialect Sentiment Analysis.}
\label{tab:results_task1}
\end{table*}

\paragraph{Analysis of Subtask 1 Results.}
Several key findings emerge from the experimental outcomes of Subtask 1:

\begin{enumerate}
    \item \textbf{Pretrained Dialectal Encoders Outperform Generative LLMs}: All top-performing submissions relied on dedicated dialectal encoder models---most notably MARBERTv2 and AraBERTv02. Models exposed to substantial dialectal social media data during pretraining consistently outperformed general Modern Standard Arabic (MSA) baselines and zero-shot prompting of large generative models.
    
    \item \textbf{Robustness via Adversarial Training and Pseudo-Labeling}: The winning system, \textbf{ALEXIS}, achieved a substantial lead (+4.57\% Macro F1 over 2nd place). This gain was primarily driven by Fast Gradient Method (FGM) adversarial training, which forces the model representations to remain invariant to dialectal surface lexical variations, paired with iterative self-training that effectively bootstrapped representations for the unlabelled Lebanese test instances.
    
    \item \textbf{Ensembling and Loss Engineering}: Teams that implemented stratified $K$-fold ensembling (\textbf{Axiom}, \textbf{faisalm93}) and model stacking across distinct pretraining families (\textbf{NU-SentEval}, \textbf{NAMAA}) demonstrated higher stability. Furthermore, incorporating Focal Loss (\textbf{NU-SentEval}) effectively addressed the subtle semantic boundaries between neutral and mildly opinionated dialectal phrases.
    
    \item \textbf{Zero-Shot Levantine Transferability}: Analysis by \textbf{PTUK-NLP} and \textbf{Lahja} showed that models trained on geographically proximate dialects (Jordanian) exhibit strong zero-shot cross-lingual transfer to Lebanese, confirming shared lexical roots within the Levantine dialect family.
\end{enumerate}

\subsection{Subtask 2: Arabic Sentiment Swap}

Table~\ref{tab:results_task2} presents the official evaluation results for Subtask 2, sorted by the primary metric, Sentiment Style Accuracy, alongside semantic preservation metrics (BLEU and chrF).

\begin{table*}[ht]
\centering
\small
\begin{tabular}{rllccc}
\toprule
\textbf{\#} & \textbf{Team Name} & \textbf{CodaBench Username} & \textbf{Sentiment Style Accuracy} & \textbf{BLEU} & \textbf{chrF} \\
\midrule
1 & \textbf{ALEXIS} & \textit{momen\_ibrahim-870527} & \textbf{0.9581} & \textbf{56.91} & 72.18 \\
2 & \textbf{AKARROUCH} & \textit{akarrouch-871040} & 0.7851 & 49.82 & 70.12 \\
3 & \textbf{LahjaSensei} & \textit{ahmadalmoustafa-865064} & 0.7789 & 41.48 & 64.45 \\
4 & \textbf{NAMAA} & \textit{mona\_-871054} & 0.7597 & 51.58 & 71.02 \\
5 & -- & \textit{melmoussaoui-870973} & 0.7530 & 49.03 & 69.53 \\
6 & \textbf{Sabaa} & \textit{mohamedsabaa-855812} & 0.7447 & 49.07 & 69.33 \\
7 & \textbf{NAMAA} & \textit{hsnhb7-868552} & 0.7343 & 46.80 & 67.10 \\
8 & \textbf{NU-SentEval} & \textit{areejkhaled-868894} & 0.6898 & 54.63 & \textbf{72.61} \\
9 & \textbf{NileNLP} & \textit{mohammed\_bahgat-863352} & 0.6732 & 41.38 & 65.84 \\
10 & \textbf{NileNLP} & \textit{ehabashraf-863355} & 0.6732 & 41.38 & 65.84 \\
11 & -- & \textit{ahmad\_alshattha-870972} & 0.6354 & 25.07 & 56.33 \\
12 & -- & \textit{hoanguitcs-868178} & 0.6220 & 49.27 & 68.26 \\
13 & \textbf{Gradient Descenders} & \textit{normalman159-868503} & 0.5997 & 11.31 & 38.22 \\
14 & -- & \textit{mustafaelkady-870032} & 0.4772 & 0.03 & 10.04 \\
15 & -- & \textit{mohamedmhaouach-872442} & 0.2807 & 43.06 & 64.61 \\
\bottomrule
\end{tabular}
\caption{Official evaluation results for Subtask 2: Arabic Sentiment Swap.}
\label{tab:results_task2}
\end{table*}

\paragraph{Analysis of Subtask 2 Results.}
Evaluating generative sentiment swapping across Arabic dialects revealed several critical insights:

\begin{enumerate}
    \item \textbf{The Polarity Transfer vs. Content Preservation Trade-off}: Generative sentiment style transfer poses an intrinsic tension between modifying sentiment markers (necessary for high Sentiment Style Accuracy) and retaining non-sentiment lexical content (measured by BLEU and chrF). Submissions that modified too few words achieved high BLEU but failed to invert sentiment (e.g., lower accuracy), while unconstrained generative rewrites suffered severe lexical drift.
    
    \item \textbf{The Impact of Preference Optimization (DPO)}: The winning team, \textbf{ALEXIS}, dominated across all dimensions (0.9581 accuracy, 56.91 BLEU, 72.18 chrF). By directly training on paired contrastive preferences using DPO, their system learned to apply minimal, surgical edits to dialectal sentiment words (e.g., negating adjectives or swapping affective idioms) while strictly preserving sentence structure.
    
    \item \textbf{Generate-and-Rerank Pipelines as an Essential Safety Mechanism}: Teams employing generate-and-rerank pipelines (\textbf{AKARROUCH}, \textbf{NAMAA}, \textbf{Sabaa}) achieved consistently higher style transfer accuracy ($>74\%$) than unguided generation. Filtering candidate rewrites with an external sentiment classifier prevented output hallucination and ensured the target polarity was reliably achieved.
    
    \item \textbf{Model Merging and PEFT for Dialect Specialization}: Advanced parameter-efficient techniques, such as SLERP model merging across dialect checkpoints (\textbf{LahjaSensei}) and Low-Rank Adaptation (\textbf{NileNLP}), proved competitive while requiring substantially lower computational footprints during fine-tuning.
\end{enumerate}

\section{Conclusion}

This overview paper presented the results and findings of the DialectSentEval shared task, held at ArabicNLP 2026. The shared task benchmarked two core dimensions of dialectal NLP: multi-class Arabic dialect sentiment classification (Subtask 1) and generative sentiment style transfer across four major Arabic dialect groups (Subtask 2). 

A total of 15 teams actively participated in the competition. The evaluation demonstrated the continued superiority of dialect-pretrained masked encoders when paired with adversarial regularization and self-training for classification, as well as the immense potential of Direct Preference Optimization (DPO) and classifier-guided reranking for controlled generative style transfer in low-resource and dialectally diverse environments. 

\bibliography{custom}

\end{document}